\documentclass[runningheads]{llncs}
\usepackage[T1]{fontenc}
\usepackage{newtxtext}
\usepackage[varvw]{newtxmath}
\usepackage{graphicx,booktabs,array}
\usepackage[hidelinks]{hyperref}
\makeatletter
\def\bibcite#1#2{\@newl@bel{b}{#1}{#2}}
\newcount\dasff@n
\newcount\dasff@i
\def\@citex[#1]#2{%
  \dasff@n\z@
  \@for\@citeb:=#2\do{\advance\dasff@n\@ne}%
  \dasff@i\z@
  \@for\@citeb:=#2\do{%
    \advance\dasff@i\@ne
    \edef\@citeb{\expandafter\@firstofone\@citeb\@empty}%
    \if@filesw\immediate\write\@auxout{\string\citation{\@citeb}}\fi
    \@ifundefined{b@\@citeb}{%
      \ifnum\dasff@i=\@ne [\fi
      \textbf{?}%
      \ifnum\dasff@i<\dasff@n ,\fi
      \ifnum\dasff@i=\dasff@n ]\fi
      \@latex@warning{Citation `\@citeb' on page \thepage \space undefined}%
    }{%
      \hyper@link{cite}{cite.\@citeb}{%
        \ifnum\dasff@i=\@ne [\fi
        \csname b@\@citeb\endcsname
        \ifnum\dasff@i<\dasff@n ,\fi
        \ifnum\dasff@i=\dasff@n ]\fi
      }%
    }%
  }%
}
\makeatother
\DeclareRobustCommand{\Figref}[1]{\hyperref[#1]{Fig.~\ref*{#1}}}
\DeclareRobustCommand{\Tabref}[1]{\hyperref[#1]{Table~\ref*{#1}}}
\DeclareRobustCommand{\Secref}[1]{\hyperref[#1]{Sect.~\ref*{#1}}}
\DeclareRobustCommand{\figpanel}[2]{\hyperref[#1]{Fig.~\ref*{#1}#2}}

\begin{document}
\title{FARE: Factor-Aware Retrieval with Error-Guided Fusion for Retail Forecasting}
\titlerunning{FARE for Retail Forecasting}
\input{author-info}
\maketitle

\begin{abstract}
Retail sales forecasting underpins inventory planning, yet promotions and calendar events reshape short-term demand.
Although each factor usually has historical occurrences, the complete factor combination of a target week often has few precedents.
The challenge is therefore to retrieve informative partial matches and weight them by their reliability, which varies across series.
We propose FARE (Factor-Aware Retrieval with Error-Guided Fusion), which ranks historical weeks separately for each known target-week factor and merges the selected weeks into a factor reference under a fixed budget.
Guided by shrunk errors from completed development weeks and the current disagreement among candidates, a first gate mixes this factor reference with a shape reference from similar sales histories, and a second gate controls how strongly the mixture corrects a backbone forecast.
Both gates remain fixed during testing.
On Favorita, M5, Rossmann, and Rohlik, FARE reduces the weighted absolute percentage error (WAPE) of five backbones by 11.46\%, 4.06\%, 14.63\%, and 10.16\% on average.
With Chronos-2 frozen, FARE also outperforms conventional and existing enhancement methods, including supervised residual correction and retrieval augmentation.

\textbf{Code} is available at \CodeRepositoryLink.
\keywords{Time series forecasting \and Factor-conditioned retrieval \and Historical prediction errors \and Retail sales forecasting}
\end{abstract}
\section{Introduction}
\begin{figure}[t]
\centering
\includegraphics[width=\textwidth]{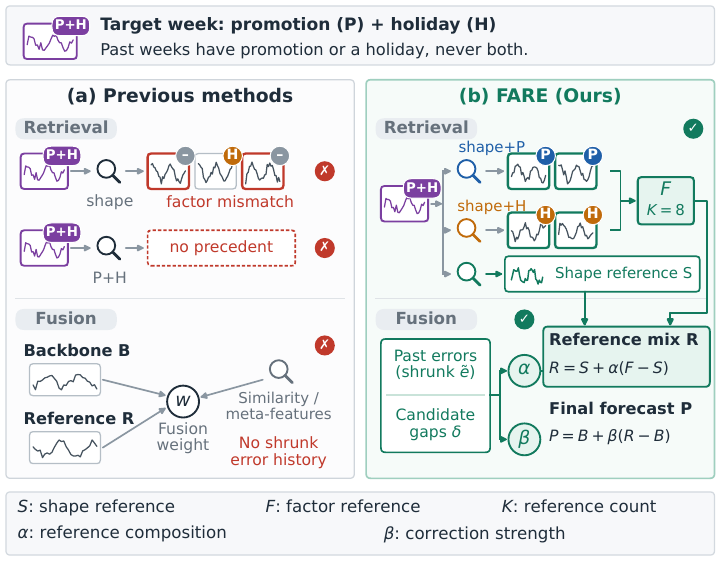}
\Description{A target week combines promotion and holiday factors. Shape-only retrieval can mismatch the factors, while requiring both factors can return no precedent. FARE retrieves separate partial factor matches and a shape reference, then uses historical errors and candidate gaps to guide reference composition and correction strength.}
\caption{Previous methods versus FARE (illustrative example). FARE retrieves partial precedents for each known target-week factor and fuses candidates with gates guided by their historical errors.}
\label{fig:motivation}
\end{figure}

Retail sales forecasts~\cite{ref28} support inventory replenishment and supply-chain planning, where errors affect product availability and cause excess stock.
Promotions and calendar events change product-level sales over short horizons, so their known future schedules are relevant to the forecast.
The difficulty sharpens when each factor has occurred historically but their complete combination in the target week has few precedents, and the forecaster must reuse observations whose factors only partially agree with the target week.

Existing methods use these signals in three ways. Covariate models such as TFT~\cite{ref10} and DD-CZFormer~\cite{ref7} condition the forecast on known factors, retrieval methods such as SARAF~\cite{ref1}, PFRP~\cite{ref2}, PIR~\cite{ref3}, and RAF~\cite{ref43} supply observed historical trajectories, and combination methods~\cite{ref4,ref6,ref30} weight candidate forecasts by past errors or current features.

When precedents for complete factor combinations are scarce, retrieval and combination face two linked challenges. First, similar past sales do not guarantee that a retrieved trajectory reflects the promotion or calendar conditions of the target week, yet requiring complete factor agreement discards useful partial precedents. Second, partial references may disagree with one another and with the backbone, and their reliability varies across series and is hard to estimate from limited error records.

We address these challenges with FARE, which couples factor-conditioned reference construction with error-guided two-level fusion. FARE ranks historical weeks separately for each target-week factor by sales-shape distance and agreement on that factor, and merges the selected weeks into a factor reference through alternating selection and deduplication. It then shrinks candidate-specific errors from completed development weeks toward a global mean and combines them with current prediction disagreements as inputs to two gates, which separately set the composition of the reference mixture and its correction strength on the backbone.

As illustrated in \Figref{fig:motivation}, FARE constructs factor references and allocates their contributions. Our contributions are threefold:
\begin{enumerate}
\item \textbf{Factor-conditioned partial references.} Separate factor rankings build an eight-window reference through alternating selection and deduplication under a fixed retrieval budget.
\item \textbf{Historical error-guided two-level fusion.} Shrunk per-candidate error histories and current disagreements guide separate gates for reference composition and correction strength, with convex weights that return continuously to the backbone.
\item \textbf{Evaluation of reference selection and allocation.} Experiments on Favorita, M5, Rossmann, and Rohlik with five backbones assess both designs through ablations and matched-candidate comparisons.
\end{enumerate}

\section{Related Work}
\label{sec:related}

\subsection{Retail Factors and Historical References}

Retail forecasting~\cite{ref28} treats promotions and calendar conditions as demand predictors, and DD-CZFormer~\cite{ref7} learns disentangled drivers for unseen factor combinations. Our setting instead targets scarce precedents for complete combinations and uses separate factor rankings to supply explicit reference trajectories. SARAF~\cite{ref1} combines temporal alignment with stationarity-based selection, PFRP~\cite{ref2} combines forecasts from a global memory and a local model, PIR~\cite{ref3} revises failed forecasts from local and global context, and Naive RAF~\cite{ref43} adds retrieved examples without fine-tuning the forecaster. RAFT~\cite{ref44} adds the futures of retrieved training segments to the input of a forecaster, and TS-RAG~\cite{ref45} mixes retrieved segments into the representations of a time series foundation model for zero-shot forecasting. Unlike these methods, FARE constructs factor-conditioned reference forecasts, learns their contribution to the backbone output, and retains explicit shape and factor references under the same history boundary and retrieval budget.

\subsection{Adaptive Forecast Combination}

Classical combination~\cite{ref4} and the review of Wang et al.~\cite{ref32} show that weighting forecasts by historical performance can help, although estimating the weights carries risks. Diebold and Pauly~\cite{ref46} reduce this risk by shrinking estimated combination weights toward equal weights.
FFORMA~\cite{ref30} and retail meta-learning~\cite{ref31} map series features to weights, and TimeFuse~\cite{ref6} learns instance-level fusion from input meta-features.
FARE instead shrinks the per-series error history of each candidate toward a global mean before it enters the gates, and its gates separate the composition of the reference forecast from its correction of the backbone.
PFRP also learns confidence and output gates, whereas FARE's two gates separately control how the two references are mixed and how much they contribute in total.

\section{Problem Definition}
\label{sec:problem}

Let $y_{i,d}\geq 0$ denote the sales of series $i$ on day $d$.
At each weekly origin $t$, the task is to predict $y_{i,t+h}$ for $h\in\{0,\ldots,6\}$ from the 56-day history
$\mathbf{x}_{i,t}=(y_{i,t-56},\ldots,y_{i,t-1})$
and the target-week factors $c$ known at the origin, such as the promotion, discount, SNAP, and calendar indicators described in \Secref{sec:splits}.

A historical reference $r=(j,u)$ pairs the 56 days preceding $u$ in series $j$ with the following seven-day trajectory.
Its target week must be fully observed before the query origin, that is, $u+6<t$, and retrieval follows the dataset-specific grouping rules in \Secref{sec:splits}.
During rolling evaluation, completed weeks enter the library under this boundary, while gate parameters and historical error statistics stay fixed.

FARE fuses three daily candidates, namely a backbone prediction $B$, a shape reference $S$ from similar sales histories, and a factor reference $F$ from separate matches on known target-week factors.
Their fusion weights depend on historical errors, current disagreements, and origin features, and the output $P_{i,t,h}$ is a nonnegative point forecast for day $t+h$.

\section{Methodology}

\begin{figure}[!t]
\centering
\includegraphics[width=\textwidth]{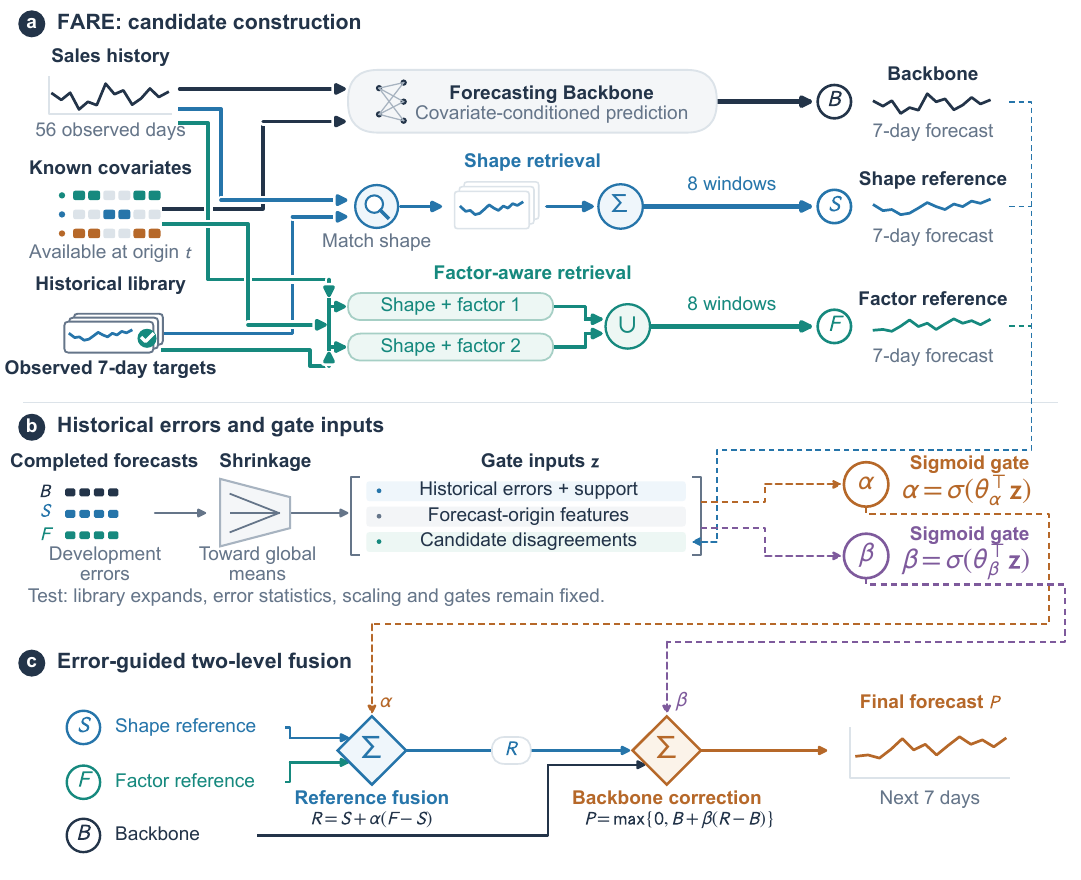}
\Description{Three panels show candidate construction, historical error inputs, and forecast allocation. FARE uses a backbone forecast B, a shape reference S, and a factor reference F as candidates. Errors from completed development forecasts are shrunk and combined with origin features and candidate disagreements. Gate alpha mixes S and F into R, and gate beta controls the correction from B toward R. During testing the library grows while error statistics and gate parameters remain fixed.}
\caption{Overview of FARE.
(a) A backbone forecast and two historical references provide three candidate forecasts.
(b) Historical candidate errors and forecast-origin features form the gating inputs.
(c) Two gates combine the references and control their contribution to the final forecast.}
\label{fig:1}
\end{figure}

As shown in \Figref{fig:1}, FARE constructs two references and fuses them in two stages. The first gate combines $S$ and $F$ into $R$, and the second sets how much $R$ adjusts $B$ to produce $P$.
Here $\operatorname{clip}(v,l,u)$ truncates $v$ to $[l,u]$, $\mathbb{I}[\cdot]$ is the indicator function, and $\sigma(v)=1/(1+\exp(-v))$.

\subsection{Three Forecast Candidates}
\label{sec:candidates}

All candidates are truncated to nonnegative values, and $B$ can come from any backbone.
The main experiments use the median forecast of the frozen Chronos-2~\cite{ref8}, with 56 days of sales and concurrent covariates as input and seven days of known future covariates, and \Secref{sec:integration} evaluates LightGBM, TFT, PatchTST, and SparseTSF as alternatives.
On all datasets, Chronos-2 receives one-hot weekdays and the two dataset-specific factors, to which retrieval matching is also restricted. Each window is scaled by its pre-origin sales level
\begin{equation}
a_{i,t}=\max\left(\frac{1}{56}\sum_{\ell=1}^{56}y_{i,t-\ell},1\right)
\label{eq:3}
\end{equation}
which reduces magnitude differences between trajectories, and the shape distance is
\begin{equation}
d_x((i,t),(j,u))= \frac{1}{56}\sum_{\ell=1}^{56} \left(\frac{y_{i,t-\ell}}{a_{i,t}}-\frac{y_{j,u-\ell}}{a_{j,u}}\right)^2
\label{eq:4}
\end{equation}
$S$ selects the $K=8$ nearest eligible windows, with ties broken by a stable sort on origin and series ID.

Favorita~\cite{ref12} encodes promotions as categories (0, 1--5, or at least 6) together with binary holidays, M5~\cite{ref11} uses binary SNAP and event indicators, Rossmann~\cite{ref34} uses binary promotions and school holidays, and Rohlik~\cite{ref35} uses binary discount and holiday indicators, where the discount indicator is one if any of its seven discount fields is positive.
For factor $q\in\{1,2\}$, the target-week mismatch is
$d_q=\frac{1}{7}\sum_{h=0}^{6} \mathbb{I}[c^q_{i,t,h}\ne c^q_{j,u,h}]$.
Each branch combines sales shape with one factor by ranking references with
$v_q=\frac{1}{2}\left(\frac{d_x}{1+d_x}+d_q\right)$, where $d_x/(1+d_x)$ puts the shape distance on a scale comparable to $d_q$.
A window can thus qualify by matching either factor, possibly for only part of the target week.
Alternating selection starts with the promotion branch on Favorita and Rossmann, the SNAP branch on M5, and the discount branch on Rohlik, and it skips duplicates so that each branch yields four windows.
Both references average their eight selected windows $\mathcal{N}_{i,t}$ with equal weights.
\begin{equation}
\widehat y^{\mathrm{ref}}_{i,t,h}= a_{i,t}\frac{1}{8}\sum_{(j,u)\in\mathcal{N}_{i,t}} \frac{y_{j,u+h}}{a_{j,u}}
\label{eq:7}
\end{equation}
$S$ and $F$ share the same eligible library and reference count.
On M5, Rossmann, and Rohlik, groups with fewer than eight eligible windows fall back to the query's previous-week sales, and the query is retained.
All references respect the pre-origin boundary, and the controlled Favorita library also excludes complete target-week combinations of high promotions and holidays. All Favorita results that use $S$ and $F$, namely FARE, its ablations, TimeFuse, and the fused backbones in \Tabref{tab:5}, use this controlled library. The exclusion removes 4,556 rows from the library before development, and 3,719 of the 311,850 test rows fall in such combinations.

\subsection{Historical Errors Accumulated Forward in Time}
\label{sec:errors}

For source $k\in\{B,S,F\}$, the normalized absolute error on an eligible completed day is
\begin{equation}
e^k_{i,u,h}=\operatorname{clip} \left(\frac{|y_{i,u+h}-\widehat y^k_{i,u,h}|}{a_{i,u}},0,10\right)
\label{eq:8}
\end{equation}
Let $\mathcal{H}_{i,t}$ hold the $n_{i,t}$ prior completed development forecast-day rows (seven per week) and $\mu^k_t$ be the mean error over all previously eligible development rows.
Errors of series with few observations are shrunk toward this global mean.
\begin{equation}
\widetilde{e}^k_{i,t}= \operatorname{clip}\left( \frac{\sum_{(u,h)\in\mathcal{H}_{i,t}}e^k_{i,u,h}+\lambda\mu^k_t} {n_{i,t}+\lambda},0,10\right),\qquad \lambda=14
\label{eq:9}
\end{equation}
The gates take the three shrunk errors and $\log(1+n)$ as inputs, and the pseudo-count $\lambda=14$ equals two weeks.
Series without history use the global mean, and errors and support are zero when no global history exists.

Development features use only origins whose seven target days precede the current origin.
Evaluation freezes the statistics of the full development period.

\subsection{Gating Features}

The base inputs are $\log(1+\text{mean sales})$ over 56 days, the history standard deviation divided by $a$ and clipped to $[0,10]$, and the zero-sales fraction.
Every dataset adds $\log(1+\text{library size})$ and its two daily factor indicators, with Favorita using the promotion term $\min(\mathrm{onpromotion},6)/6$ in place of a binary promotion flag.
Favorita, Rossmann, and Rohlik also add the branch-wise $\log(1+\text{mean ranking score})$ over the four references selected by each branch, which is set to zero on Rossmann and Rohlik when reference construction falls back to previous-week sales.
On Favorita, the library size enters once per branch, and the two values coincide because both branches search the same library.

Historical errors describe past accuracy, and current disagreements measure the size of candidate corrections.
\begin{equation}
\begin{aligned}
\delta^{BS}_{i,t,h}
&=\operatorname{clip}\left(\frac{|B_{i,t,h}-S_{i,t,h}|}{a_{i,t}},0,10\right),\\
\delta^{FS}_{i,t,h}
&=\operatorname{clip}\left(\frac{|F_{i,t,h}-S_{i,t,h}|}{a_{i,t}},0,10\right).
\end{aligned}
\label{eq:10}
\end{equation}
$\mathbf{z}$ stacks the base and factor inputs, the two disagreements, and the four historical inputs.
Binary indicators keep their values, other continuous inputs are standardized on gating training rows (standard deviation floor $10^{-8}$), and an intercept is added.
The complete models have 15, 12, 14, and 14 non-intercept inputs on Favorita, M5, Rossmann, and Rohlik, and they are fitted separately for each dataset and backbone.

\subsection{Two Bounded Soft Gates}

The gate weights are
\begin{equation}
\alpha_{i,t,h}=\sigma(\boldsymbol{\theta}_\alpha^\top\mathbf{z}_{i,t,h}), \qquad \beta_{i,t,h}=\sigma(\boldsymbol{\theta}_\beta^\top\mathbf{z}_{i,t,h})
\label{eq:11}
\end{equation}
With a shared input and separate parameters, the first gate constructs
\begin{equation}
R_{i,t,h}=S_{i,t,h}+\alpha_{i,t,h}(F_{i,t,h}-S_{i,t,h})
\label{eq:12}
\end{equation}
and the second produces
\begin{equation}
P_{i,t,h}=\max\{0,B_{i,t,h}+\beta_{i,t,h}(R_{i,t,h}-B_{i,t,h})\}
\label{eq:13}
\end{equation}
The effective weights of $B$, $S$, and $F$ are therefore $1-\beta$, $\beta(1-\alpha)$, and $\beta\alpha$.
They are nonnegative and sum to one, so $P$ lies in the convex hull of the three candidates and returns continuously to $B$ as $\beta$ approaches zero.
The hierarchy thus separates reference composition ($\alpha$) from total reference participation ($\beta$), which \Secref{sec:mechanism-analysis} evaluates with matched-candidate controls.

\subsection{Fitting Objective and Optimization}
\label{sec:optimization}

The gates are fitted on development rows $\mathcal{D}_{\mathrm{dev}}$ by minimizing
\begin{equation}
\begin{split}
\mathcal{L}
&= \frac{
\sum_{(i,t,h)\in\mathcal{D}_{\mathrm{dev}}}
\sqrt{(P_{i,t,h}-y_{i,t+h})^2+1}
}{
\sum_{(i,t,h)\in\mathcal{D}_{\mathrm{dev}}}y_{i,t+h}
}\\
&\quad + 0.001\left(
\|\boldsymbol{\theta}_{\alpha,\mathrm{nonbias}}\|_2^2
+\|\boldsymbol{\theta}_{\beta,\mathrm{nonbias}}\|_2^2
\right).
\end{split}
\label{eq:14}
\end{equation}
The first term is a smooth surrogate of normalized absolute error with a smoothing constant of one sales unit. The second term regularizes the non-intercept gate parameters, and evaluation uses unsmoothed WAPE and MAE.
Zero-initialized parameters are optimized by L-BFGS-B with analytical gradients (at most 250 iterations, $\mathrm{ftol}=10^{-9}$, $\mathrm{gtol}=10^{-6}$).
Gate fitting and error accumulation both use development labels under the boundary in \Secref{sec:problem}.

\section{Experimental Setup}

\subsection{Data and Temporal Protocol}
\label{sec:splits}

\textbf{Favorita}~\cite{ref12} has 1,782 store-category series with promotion and holiday factors, and its references are grouped by product category. Missing Christmas records are not zero-filled, so origins whose 56-day history or target week includes 25 December are skipped, which removes nine origins from each holiday season.
\textbf{M5}~\cite{ref11} has 30,490 item-store series with SNAP and calendar-event factors, grouped by item. A series enters at its first week with a listed price and needs 56 days of coverage, but price values are not used. M5 series are mostly intermittent, with zero sales on 62\% of history days in the average development row, and their WAPE is correspondingly high.
\textbf{Rossmann}~\cite{ref34} has 1,115 store series with promotion and school-holiday factors, grouped by store type and assortment. Evaluation covers all days, including 162,990 open-day rows, without forcing closed-day forecasts to zero.
\textbf{Rohlik V2}~\cite{ref35} has 5,390 anonymized product series of availability-adjusted sales in item counts or kilograms, with discount and holiday indicators, grouped by warehouse and level-2 category.
On all datasets, each weekly origin uses 56 complete calendar days to forecast the next seven nonoverlapping days, incomplete histories are skipped, and our rolling protocol replaces the original competition horizons and scoring rules.

The fitting, development, and testing periods are listed in \Tabref{tab:1}.
Chronos-2~\cite{ref8} stays frozen in all comparisons, whereas LightGBM~\cite{ref9}, TFT~\cite{ref10}, PatchTST~\cite{ref37}, and SparseTSF~\cite{ref38} are trained on the fitting data, with configurations selected on fitting-period validation origins before retraining on all fitting origins.
Correction modules are selected by chronological splits of the development period, whose labels fit FARE's gates and the LightGBM residual corrector.
Test labels are never used for fitting, selection, or error calibration, although completed test weeks may supply history for later origins.

\begin{table}[!htbp]
\centering\small
\caption{Temporal ranges and forecast-day row counts. Fit denotes supervised-backbone fitting and historical references, and Dev. denotes correction-module development.}
\label{tab:1}
\setlength{\tabcolsep}{4pt}
\begin{tabular*}{\textwidth}{@{\extracolsep{\fill}}llrlr@{}}
\toprule
\textbf{Dataset} & \textbf{Stage} & \textbf{Origins} & \textbf{Origin range} & \textbf{Forecast-day rows} \\
\midrule
Favorita & Fit & 62 & 2015-02-23 to 2016-06-27 & 773,388 \\
Favorita & Dev. & 24 & 2016-07-04 to 2016-12-12 & 299,376 \\
Favorita & Test & 25 & 2017-02-20 to 2017-08-07 & 311,850 \\
M5 & Fit & 165 & 2011-03-28 to 2014-05-19 & 23,520,098 \\
M5 & Dev. & 52 & 2014-05-26 to 2015-05-18 & 10,539,837 \\
M5 & Test & 52 & 2015-05-25 to 2016-05-16 & 11,053,322 \\
\addlinespace[2pt]
Rossmann & Fit & 73 & 2013-03-04 to 2014-07-21 & 564,725 \\
Rossmann & Dev. & 26 & 2014-07-28 to 2015-01-19 & 170,170 \\
Rossmann & Test & 26 & 2015-01-26 to 2015-07-20 & 196,630 \\
\addlinespace[2pt]
Rohlik & Fit & 140 & 2020-09-28 to 2023-05-29 & 1,300,544 \\
Rohlik & Dev. & 26 & 2023-06-05 to 2023-11-27 & 321,573 \\
Rohlik & Test & 26 & 2023-12-04 to 2024-05-27 & 283,185 \\
\bottomrule
\end{tabular*}
\end{table}

\subsection{Comparisons and Controls}
\label{sec:extensions}

Except for the backbone study in \Secref{sec:integration}, all comparisons and ablations use frozen Chronos-2 as the backbone of FARE, and all contrasts use the evaluation rows of the complete model in each dataset.
Input ablations remove the historical errors and support, the disagreements, or the factor branch with its inputs.
Component controls substitute joint Top-8 retrieval, three-seed random retrieval, global errors (retaining series-specific support), unshrunk errors, or a learned constant gate.
Output-path controls keep all gate inputs. Learned B+S fixes $\alpha=0$ and learned B+F fixes $\alpha=1$, both learning $\beta$ on development data with $k$ parameters versus $2k$ in the complete model, while the joint-constant control learns only the two gate intercepts.

TimeFuse~\cite{ref6} combines our three stored candidates through meta-feature weighting.
PFRP~\cite{ref2} and its DLinear~\cite{ref36} backbone use sales history, PIR~\cite{ref3} revises frozen Chronos-2 forecasts with deterministic calendar features and cross-series retrieval over fitting windows, and SARAF~\cite{ref1} uses a fitted linear predictor and same-series history.
Because these adapted systems differ in backbones, covariates, and libraries, matched-candidate controls and within-backbone ablations assess allocation more directly. PFRP, PIR, and SARAF report means over three seeds.

RAF and LightGBM residual correction share the frozen Chronos-2 weights, known covariates, and evaluation rows.
Our retail adaptation of Naive RAF~\cite{ref43} uses Chronos-2 for embeddings and forecasts with a weekly fitting-only library.
It retrieves one 56+7-day example from the legal fitting-period windows of the same series by the summed L2 distance of corresponding tokens, and series without such windows use up to 64 fixed global examples from the last fitting origin, evenly spaced in series-ID order.
The example and the 56-day query are normalized separately, and the example is shifted to join the query continuously, giving a 119-day input. Source-date covariates are retained, and forecasts are restored to the query scale.

The LightGBM~\cite{ref9} residual corrector learns $y-B$ from 56-day lags, historical summaries, forecast step, date features, permitted future factors, and the clipped frozen forecast $B$, returning $\max(B+\hat r,0)$.
Using only development labels, it selects the number of leaves in $\{15,31,63\}$ by corrected-forecast pooled WAPE on the last $\lceil0.2T_{\mathrm{dev}}\rceil$ origins after fitting on earlier ones, trains 300 L1 boosting rounds, and refits 300 rounds on all development origins.
We report the mean of three separately scored seeds, not an ensemble.
This corrector differs from the LightGBM backbone in \Tabref{tab:5}, and Chronos-2 is neither retrained nor fine-tuned.

The conventional baselines are ETS~\cite{ref39}, SARIMA~\cite{ref40}, Ridge regression~\cite{ref41}, and XGBoost~\cite{ref42}.
ETS and SARIMA are refit on the latest 56 days at each origin, while Ridge and XGBoost also use permitted future factors and deterministic date features.
Settings are selected on the last eight fitting origins, after which Ridge and XGBoost are retrained on all fitting origins.
ETS selects AAA on Favorita and Rossmann and ANN on M5 and Rohlik, SARIMA uses $(1,0,0)\times(0,1,1)_7$, Ridge uses a regularization strength of $10{,}000$, and XGBoost uses depth 7, a learning rate of 0.05, and 300 rounds.
ETS, SARIMA, and Ridge report one run, and XGBoost reports the mean of three separately scored seeds.

For supervised backbones, candidate errors and disagreements are recomputed and the gates are refit for each backbone, and LightGBM, TFT, PatchTST, and SparseTSF results are means over three seeds on all datasets.
PatchTST~\cite{ref37} and SparseTSF~\cite{ref38} use only sales history with the same 56-day input and seven-day horizon, and their settings are selected on the last eight fitting origins before retraining.

\subsection{Metrics and Implementation}

For evaluation rows $\mathcal{E}$, the two metrics are defined as
\begin{equation}
\mathrm{WAPE}=\frac{\sum_{r\in\mathcal{E}}|y_r-P_r|}{\sum_{r\in\mathcal{E}}y_r}, \qquad \mathrm{MAE}=\frac{\sum_{r\in\mathcal{E}}|y_r-P_r|}{|\mathcal{E}|}
\label{eq:15}
\end{equation}
WAPE pools absolute errors and sales across weeks, rather than averaging weekly values, and is reported in percent.
MAE is in daily-sales units and equals WAPE times mean sales on the same targets.

Chronos-2 uses the pretrained \texttt{amazon/chronos-2} model with nonnegative median forecasts and cross-series learning disabled. Each backbone forecast therefore depends only on the history and known factors of its own series, which keeps the forecasts independent of batch composition and leaves retrieval as the only channel through which other series inform FARE.
FARE uses sales-shape distances, whereas the RAF adaptation reuses frozen Chronos-2 embeddings.
Reference aggregation costs $O(KH)$ after selection. Search and sorting, however, scale with library size, which the parameter count does not reflect. On M5, building the references for all 104 development and test origins takes 289 s, and fitting and applying the gates for Chronos-2, LightGBM, and three TFT seeds takes 1,141 s.
Including intercepts, the two gates have 32, 26, 30, and 30 trainable parameters on Favorita, M5, Rossmann, and Rohlik, optimized as in \Secref{sec:optimization}.
All supervised-backbone and fusion runs use an NVIDIA RTX 4090.

\section{Results and Analysis}

\subsection{Comparison with Existing Methods}
\label{sec:system-comparison}

As reported in \Tabref{tab:2}, FARE with Chronos-2 achieves the lowest WAPE and MAE on all four datasets.
On Favorita and Rohlik, the strongest competitor is TimeFuse, which fuses the same three candidates, and FARE lowers its WAPE by a further 6.81\% and 6.69\%, showing that FARE's allocation draws more from the same candidates.
On M5 and Rossmann, FARE still beats the strongest competitors, LightGBM residual correction and XGBoost, which use the factors directly, by 3.64\% and 1.32\%, while keeping Chronos-2 frozen and fitting only 26 to 30 gate parameters.

\begin{table}[t]
\centering\fontsize{8}{9.5}\selectfont
\caption{Comparison with conventional and conference methods. FARE uses frozen Chronos-2 as its backbone, and TimeFuse, PIR, RAF, and LightGBM residual correction build on the same frozen forecasts.}
\label{tab:2}
\setlength{\tabcolsep}{1pt}
\begin{tabular}{@{}l@{\hspace{4pt}}rrrrrrrr@{}}
\toprule
\raisebox{-1.6ex}[0pt][0pt]{Method} & \multicolumn{2}{c}{\textbf{Favorita}} & \multicolumn{2}{c}{\textbf{M5}} & \multicolumn{2}{c}{\textbf{Rossmann}} & \multicolumn{2}{c}{\textbf{Rohlik}} \\
\cmidrule(lr){2-3}\cmidrule(lr){4-5}\cmidrule(lr){6-7}\cmidrule(lr){8-9}
 & WAPE$\downarrow$ & MAE$\downarrow$ & WAPE$\downarrow$ & MAE$\downarrow$ & WAPE$\downarrow$ & MAE$\downarrow$ & WAPE$\downarrow$ & MAE$\downarrow$ \\
\midrule
\multicolumn{9}{@{}l}{\textit{Conventional methods}} \\
ETS~\cite{ref39} & 17.72 & 85.4 & 75.47 & 0.992 & 25.07 & 1477 & 27.00 & 37.1 \\
SARIMA~\cite{ref40} & 15.94 & 76.8 & 77.85 & 1.023 & 23.22 & 1368 & 26.39 & 36.3 \\
Ridge~\cite{ref41} & 18.58 & 89.6 & 74.99 & 0.986 & 18.52 & 1091 & 25.80 & 35.4 \\
XGBoost~\cite{ref42} & 15.63 & 75.4 & 73.30 & 0.964 & 15.91 & 937 & 20.92 & 28.7 \\
LightGBM residual~\cite{ref9} & 15.15 & 73.0 & 66.54 & 0.875 & 17.94 & 1057 & 20.10 & 27.6 \\
\midrule
\multicolumn{9}{@{}l}{\textit{Conference methods}} \\
(ICML 25') TimeFuse~\cite{ref6} & 14.25 & 68.7 & 67.48 & 0.887 & 18.74 & 1104 & 19.87 & 27.3 \\
(AAAI 26') PFRP~\cite{ref2} & 17.03 & 82.1 & 73.44 & 0.965 & 21.23 & 1251 & 22.88 & 31.4 \\
(NeurIPS 25') PIR~\cite{ref3} & 15.58 & 75.1 & 72.96 & 0.959 & 20.53 & 1210 & 20.23 & 27.8 \\
(KDD 26') SARAF~\cite{ref1} & 21.60 & 104.1 & 73.94 & 0.972 & 19.72 & 1162 & 25.64 & 35.2 \\
(AISTATS 26') RAF~\cite{ref43} & 14.80 & 71.4 & 68.70 & 0.903 & 20.88 & 1230 & 20.58 & 28.3 \\
\midrule
\textbf{FARE (Ours)} & \textbf{13.28} & \textbf{64.0} & \textbf{64.12} & \textbf{0.843} & \textbf{15.70} & \textbf{925} & \textbf{18.54} & \textbf{25.5} \\
\bottomrule
\end{tabular}
\end{table}

With the same frozen Chronos-2, FARE reduces the backbone WAPE reported in \Tabref{tab:5} by 12.29\%, 6.13\%, 25.94\%, and 9.96\% on Favorita, M5, Rossmann, and Rohlik, compared with at most 5.88\%, 2.59\%, 15.38\%, and 3.50\% for the other correction modules, and its advantage is largest on Rossmann.
Retrieval augmentation and residual correction each fall below the backbone on one dataset, whereas FARE improves it on all four.

\subsection{Ablation Study}
\label{sec:ablations}

As summarized in \Tabref{tab:ablations}, removing the historical errors, the disagreements, or the factor branch increases WAPE on all four datasets, and every component therefore contributes. Random references are the most damaging variant on every dataset, which makes the retrieved content the foundation of the correction, but the bottleneck beyond it differs by dataset. On Rossmann, removing the factor branch raises WAPE by 18.66\%, more than any control except random references, and supplying factor-matched precedents is the bottleneck there. On Favorita, fixing $\alpha$ or $\beta$ or using global errors raises WAPE by 15.14--16.49\%, about twice the 7.61--8.21\% of removing any single input, which points to candidate weighting as the bottleneck. On M5 and Rohlik, error estimation and retrieval matter more than either gate alone, since global or unshrunk errors and joint Top-8 retrieval raise WAPE by 7.80--9.19\% and 8.63--9.98\%, whereas fixing $\alpha$ or $\beta$ costs at most 4.43\% and 4.53\%. Because the bottleneck changes across datasets, FARE keeps both levels, and the complete model outperforms every variant on all four datasets. Under the same eight-reference budget, factor-wise ranking also beats joint Top-8 retrieval on every dataset, because reserving references for each factor keeps partial precedents that a combined score would rank below windows with similar shape but mismatched factors.

With all gate inputs retained, learned B+F beats learned B+S on all four datasets, which makes $F$ the more informative correction target, and both beat the joint-constant control, while none of the three matches the complete FARE. A fitted $\beta$ does not by itself protect the backbone, as seven variants on Favorita, five on M5, and two on Rohlik fall below it. Fitted on development rows, $\beta$ stays small on test rows only when its inputs separate reliable from unreliable references, and changed references, pooled or unshrunk errors, or restricted gates weaken this signal, much as the development-fitted global weights degrade M5 in \figpanel{fig:mechanism-summary}{a}.

\begin{table}[!t]
\centering\fontsize{8}{9.5}\selectfont
\caption{FARE component ablations with Chronos-2.}
\label{tab:ablations}
\setlength{\tabcolsep}{1pt}
\begin{tabular}{@{}l@{\hspace{4pt}}rrrrrrrr@{}}
\toprule
\raisebox{-1.6ex}[0pt][0pt]{Variant} & \multicolumn{2}{c}{Favorita} & \multicolumn{2}{c}{M5} & \multicolumn{2}{c}{Rossmann} & \multicolumn{2}{c}{Rohlik} \\
\cmidrule(lr){2-3}\cmidrule(lr){4-5}\cmidrule(lr){6-7}\cmidrule(lr){8-9}
 & WAPE$\downarrow$ & MAE$\downarrow$ & WAPE$\downarrow$ & MAE$\downarrow$ & WAPE$\downarrow$ & MAE$\downarrow$ & WAPE$\downarrow$ & MAE$\downarrow$ \\
\midrule
Chronos-2 backbone & 15.14 & 73.0 & 68.31 & 0.898 & 21.20 & 1249 & 20.59 & 28.3 \\
\textbf{FARE (Ours)} & \textbf{13.28} & \textbf{64.0} & \textbf{64.12} & \textbf{0.843} & \textbf{15.70} & \textbf{925} & \textbf{18.54} & \textbf{25.5} \\
\midrule
\multicolumn{9}{l}{\textit{Input and reference-branch ablations}} \\
w/o historical errors & 14.34 & 69.1 & 66.97 & 0.880 & 16.86 & 994 & 19.65 & 27.0 \\
w/o factor reference & 14.37 & 69.3 & 66.98 & 0.880 & 18.63 & 1098 & 20.11 & 27.6 \\
w/o disagreements & 14.29 & 68.9 & 67.41 & 0.886 & 18.00 & 1061 & 19.65 & 27.0 \\
\midrule
\multicolumn{9}{l}{\textit{Retrieval, error estimation, and gate ablations}} \\
Constant $\alpha$ & 15.29 & 73.7 & 66.54 & 0.875 & 16.60 & 978 & 19.14 & 26.3 \\
Constant $\beta$ & 15.33 & 73.9 & 66.96 & 0.880 & 16.92 & 997 & 19.38 & 26.6 \\
Global errors & 15.47 & 74.6 & 69.61 & 0.915 & 17.81 & 1050 & 20.14 & 27.7 \\
w/o shrinkage & 14.88 & 71.7 & 70.01 & 0.920 & 18.28 & 1077 & 20.35 & 28.0 \\
Joint Top-8 & 14.97 & 72.2 & 69.12 & 0.909 & 17.20 & 1014 & 20.39 & 28.0 \\
Random references & 16.94 & 81.6 & 71.82 & 0.944 & 19.31 & 1138 & 22.16 & 30.4 \\
\midrule
\multicolumn{9}{l}{\textit{Candidate-path and joint-constant controls}} \\
Learned B+S & 15.78 & 76.0 & 67.75 & 0.891 & 17.44 & 1028 & 20.10 & 27.6 \\
Learned B+F & 15.62 & 75.3 & 67.09 & 0.882 & 17.20 & 1014 & 19.94 & 27.4 \\
Constant $\alpha$ and $\beta$ & 16.14 & 77.8 & 68.46 & 0.900 & 18.11 & 1067 & 21.14 & 29.0 \\
\bottomrule
\end{tabular}
\end{table}

\subsection{Mechanism Analysis of Reference Provision versus Allocation}
\label{sec:mechanism-analysis}

\begin{figure}[t]
\centering
\setlength{\abovecaptionskip}{4pt}
\includegraphics[width=0.78\textwidth]{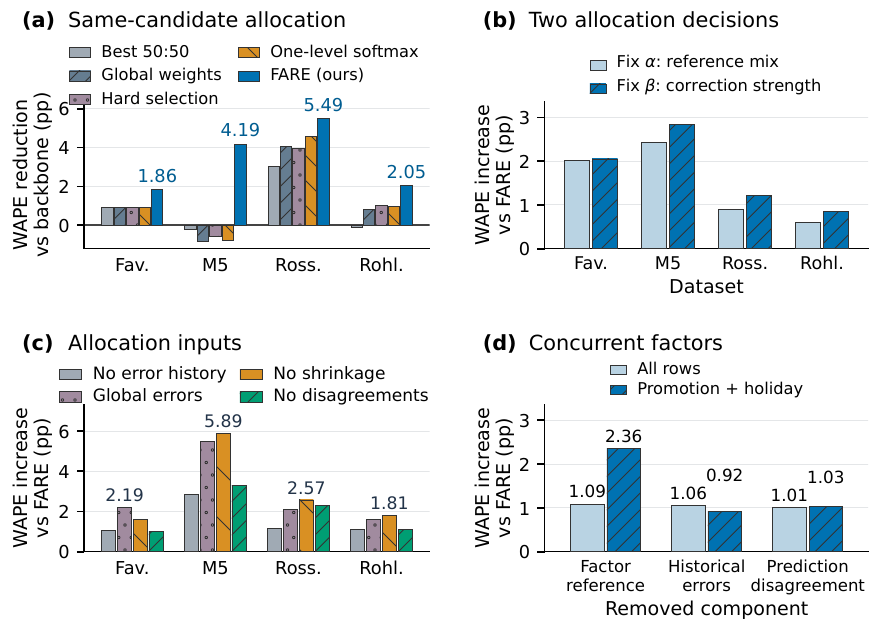}
\Description{Four grouped-bar panels compare forecast allocation and component effects. Panel a shows WAPE reductions from alternative rules applied to the same candidates. Panel b contrasts fixing reference composition and fixing correction strength across four datasets. Panel c shows error increases after removing or replacing allocation inputs. Panel d contrasts component effects on all Favorita rows and the promotion-holiday subgroup. Numeric labels and patterns supplement the color encoding.}
\caption{Reference allocation and component effects in FARE on the final test rows. (a) WAPE reduction from the backbone for alternative combination rules. Best 50:50 reports the lower test WAPE of the fixed 50:50 B/S and B/F blends. (b) Effects of fixing reference composition ($\alpha$) or correction strength ($\beta$). (c) WAPE increase over full FARE after changing allocation inputs. (d) Component effects on all Favorita rows and the promotion--holiday subgroup.}
\label{fig:mechanism-summary}
\end{figure}

As illustrated in \figpanel{fig:mechanism-summary}{a}, the one-level softmax isolates the two-level structure because it shares the candidates, inputs, loss, and parameter count of FARE. On M5, it degrades the backbone, as do fixed blends, global weights, and historical hard selection, whereas FARE improves it. Convexity alone therefore does not prevent degradation. With the same candidates, inputs, loss, and parameter count, the remaining difference is that FARE separates what to mix ($\alpha$) from how far to move ($\beta$). As shown in \figpanel{fig:mechanism-summary}{b}, fixing either gate raises WAPE on all four datasets, which supports learning the two decisions separately.

The error history helps only when it is series-specific and shrunk. On all four datasets, global or unshrunk errors are worse than no error history. An inaccurate reliability estimate thus misguides the gates, and shrinkage makes series-level estimates usable from the few development weeks that scarce combinations provide. Disagreements complement this history. As shown in \figpanel{fig:mechanism-summary}{c}, removing them hurts every dataset even with the error history and origin features intact, since past errors describe how reliable a candidate has been, whereas disagreements describe how large the current correction would be. Finally, as shown in \figpanel{fig:mechanism-summary}{d}, removing the factor branch on Favorita raises WAPE by 2.36 points in the high-promotion and holiday subgroup but by 1.09 points over all test rows, and partial factor precedents therefore matter more where the complete combination is rare.

\subsection{Complementarity Across Forecasting Backbones}
\label{sec:integration}

\begin{table}[!htbp]
\centering\fontsize{8}{9.5}\selectfont
\caption{Backbone forecasts with and without FARE.}
\label{tab:5}
\setlength{\tabcolsep}{0.8pt}
\begin{tabular}{@{}l@{\hspace{0pt}}lrrrrrrrr@{}}
\toprule
\raisebox{-1.6ex}[0pt][0pt]{Backbone} & \raisebox{-1.6ex}[0pt][0pt]{FARE} & \multicolumn{2}{c}{Favorita} & \multicolumn{2}{c}{M5} & \multicolumn{2}{c}{Rossmann} & \multicolumn{2}{c}{Rohlik} \\
\cmidrule(lr){3-4}\cmidrule(lr){5-6}\cmidrule(lr){7-8}\cmidrule(lr){9-10}
 & & WAPE$\downarrow$ & MAE$\downarrow$ & WAPE$\downarrow$ & MAE$\downarrow$ & WAPE$\downarrow$ & MAE$\downarrow$ & WAPE$\downarrow$ & MAE$\downarrow$ \\
\midrule
(arXiv 25') Chronos-2~\cite{ref8} & without & 15.14 & 73.0 & 68.31 & 0.898 & 21.20 & 1249 & 20.59 & 28.3 \\
 & \textbf{with} & \textbf{13.28} & \textbf{64.0} & \textbf{64.12} & \textbf{0.843} & \textbf{15.70} & \textbf{925} & \textbf{18.54} & \textbf{25.5}\\
\midrule
(NeurIPS 17') LightGBM~\cite{ref9} & without & 13.70 & 66.0 & 65.78 & 0.865 & 13.01 & 767 & 18.60 & 25.5 \\
 & \textbf{with} & \textbf{11.98} & \textbf{57.7} & \textbf{63.01} & \textbf{0.828} & \textbf{11.98} & \textbf{706} & \textbf{16.49} & \textbf{22.7} \\
\midrule
(IJF 21') TFT~\cite{ref10} & without & 13.91 & 67.1 & 64.83 & 0.852 & 14.57 & 859 & 18.48 & 25.4 \\
 & \textbf{with} & \textbf{12.43} & \textbf{59.9} & \textbf{62.57} & \textbf{0.823} & \textbf{12.80} & \textbf{754} & \textbf{16.23} & \textbf{22.3} \\
\midrule
(ICLR 23') PatchTST~\cite{ref37} & without & 14.50 & 69.9 & 66.18 & 0.870 & 17.73 & 1045 & 21.21 & 29.1 \\
 & \textbf{with} & \textbf{13.16} & \textbf{63.5} & \textbf{63.24} & \textbf{0.831} & \textbf{15.73} & \textbf{927} & \textbf{19.29} & \textbf{26.5} \\
\midrule
(ICML 24') SparseTSF~\cite{ref38} & without & 15.37 & 74.1 & 74.03 & 0.973 & 18.79 & 1107 & 23.22 & 31.9 \\
 & \textbf{with} & \textbf{13.44} & \textbf{64.8} & \textbf{72.53} & \textbf{0.953} & \textbf{15.81} & \textbf{932} & \textbf{21.30} & \textbf{29.3} \\
\bottomrule
\end{tabular}
\end{table}

As reported in \Tabref{tab:5}, FARE improves all five backbones on all four datasets, reducing WAPE by 11.46\%, 4.06\%, 14.63\%, and 10.16\% on average. Each of the 48 fused supervised-backbone runs, across four datasets and three seeds, has lower WAPE than its own unfused backbone, and the largest seed standard deviation of the fused WAPE is 0.87 points. The gain depends more on the dataset than on the backbone. On Favorita and Rohlik, the reductions stay within 9.24--12.56\% and 8.27--12.18\% across pretrained, gradient-boosted, and deep architectures, indicating that FARE corrects an error component these backbones share. On Rossmann, FARE narrows the WAPE range across backbones from 8.19 to 3.83 points, and the weakest backbone, Chronos-2, gains the most at 25.94\%, or 23.90\% when the closed days with zero sales are excluded. Because Chronos-2 already receives the target-week factors, this gain indicates that retrieved trajectories add realized sales shapes beyond covariate access, while the sales-only PatchTST and SparseTSF additionally benefit from the factor information that FARE introduces. A strong backbone still matters, as fused SparseTSF trails unfused Chronos-2 on M5.

\subsection{Temporal Consistency and Remaining Regressions}

The gain of FARE is consistent over time. With Chronos-2, it improves 22 of 25 Favorita origins, every M5 and Rohlik origin, and 24 of 26 Rossmann origins, including Rossmann's open days, so the gain is not confined to promotion or holiday weeks. With LightGBM and TFT, every seed improves at all Rohlik origins.

FARE trails the backbone at three Favorita and two Rossmann origins, yet it beats the one-level softmax, which uses the same convex form, at every Rossmann and Rohlik origin, so the two-level structure sustains the gain.

\section{Discussion and Limitations}

The results point to two design choices for retail retrieval. Known target-week factors belong in the retrieval key, because series with similar histories can face different promotions and holidays. Reliability weights work best when per-series error estimates are shrunk toward a pooled mean, and the retrieval and allocation ablations in \Secref{sec:ablations} show which level to improve first on a new dataset. FARE adds 26 to 32 parameters to a frozen backbone and can be attached without retraining it, although its gates are refit for each backbone.

The evidence covers two known factors and seven-day forecasts on four retail datasets. Favorita's controlled library tests the reuse of partial precedents, but generalization to combinations absent from all training data remains open, and evaluating FARE across levels of combination support would show when partial retrieval helps most. Error statistics and gate parameters stay fixed during testing, yet the average gain per origin is similar in the first and second halves of the test period, at 4.19 and 4.13 points on M5, 4.97 and 6.59 points on Rossmann, and 2.38 and 1.90 points on Rohlik. When all candidates miss the same demand change, convex fusion cannot recover it, and richer candidate forecasts are one remedy.

\section{Conclusion}
To address retail forecasting when the complete factor combination of a target week has few historical precedents, this paper proposes FARE, which retrieves partial precedents separately for each known target-week factor and then fuses them with the backbone forecast through error-guided two-level gates.
Across all 20 pairings of five backbones and four benchmarks, FARE lowers both WAPE and MAE, with average relative WAPE reductions of 11.46\%, 4.06\%, 14.63\%, and 10.16\% on Favorita, M5, Rossmann, and Rohlik, while adding only a few dozen parameters to a frozen backbone.
The experiments support three conclusions.
First, retrieval should be conditioned on known future conditions rather than on historical similarity alone.
Second, under scarce evidence, candidate reliability should be estimated per series and shrunk toward a pooled mean, in the spirit of empirical Bayes.
Third, separating reference composition from correction strength outperforms one-level and fixed allocation of the same candidates.

\begin{credits}
\subsubsection{\discintname}
The authors declare no competing interests or financial conflicts.
\end{credits}

\bibliographystyle{splncs04}
\bibliography{references}

\begin{thebibliography}{10}
\providecommand{\url}[1]{\texttt{#1}}
\providecommand{\urlprefix}{URL }
\providecommand{\doi}[1]{https://doi.org/#1}

\bibitem{ref8}
Ansari, A.F., Shchur, O., K{\"u}ken, J., Auer, A., Han, B., Mercado, P., et~al.: {Chronos-2}: From univariate to universal forecasting. arXiv preprint arXiv:2510.15821 (2025). \doi{10.48550/arXiv.2510.15821}

\bibitem{ref4}
Bates, J.M., Granger, C.W.J.: The combination of forecasts. Operational Research Quarterly  \textbf{20}(4),  451--468 (1969). \doi{10.1057/jors.1969.103}

\bibitem{ref40}
Box, G.E.P., Jenkins, G.M., Reinsel, G.C., Ljung, G.M.: Time Series Analysis: Forecasting and Control. Wiley, Hoboken, NJ, 5th edn. (2015)

\bibitem{ref42}
Chen, T., Guestrin, C.: {XGBoost}: A scalable tree boosting system. In: Proceedings of the 22nd ACM SIGKDD International Conference on Knowledge Discovery and Data Mining. pp. 785--794. Association for Computing Machinery (2016). \doi{10.1145/2939672.2939785}

\bibitem{ref12}
Cook, A., {DanB}, {inversion}, Holbrook, R.: Store sales - time series forecasting. Kaggle (2021), \url{https://www.kaggle.com/competitions/store-sales-time-series-forecasting}, dataset. Accessed 20 October 2026

\bibitem{ref46}
Diebold, F.X., Pauly, P.: The use of prior information in forecast combination. International Journal of Forecasting  \textbf{6},  503--508 (1990)

\bibitem{ref2}
Du, D., Han, T., Guo, S.: Predicting the future by retrieving the past. Proceedings of the AAAI Conference on Artificial Intelligence  \textbf{40}(25),  20896--20904 (2026). \doi{10.1609/aaai.v40i25.39230}

\bibitem{ref28}
Fildes, R., Ma, S., Kolassa, S.: Retail forecasting: Research and practice. International Journal of Forecasting  \textbf{38}(4),  1283--1318 (2022). \doi{10.1016/j.ijforecast.2019.06.004}

\bibitem{ref34}
{FlorianKnauer}, Cukierski, W.: Rossmann store sales. Kaggle (2015), \url{https://www.kaggle.com/competitions/rossmann-store-sales}, dataset. Accessed 20 October 2026

\bibitem{ref44}
Han, S., Lee, S., Cha, M., Arik, S.O., Yoon, J.: Retrieval augmented time series forecasting. In: Proceedings of the 42nd International Conference on Machine Learning. Proceedings of Machine Learning Research, vol.~267, pp. 21774--21797. PMLR (2025), \url{https://proceedings.mlr.press/v267/han25d.html}

\bibitem{ref41}
Hoerl, A.E., Kennard, R.W.: Ridge regression: Biased estimation for nonorthogonal problems. Technometrics  \textbf{12}(1),  55--67 (1970). \doi{10.1080/00401706.1970.10488634}

\bibitem{ref39}
Hyndman, R.J., Koehler, A.B., Snyder, R.D., Grose, S.: A state space framework for automatic forecasting using exponential smoothing methods. International Journal of Forecasting  \textbf{18}(3),  439--454 (2002). \doi{10.1016/S0169-2070(01)00110-8}

\bibitem{ref9}
Ke, G., Meng, Q., Finley, T., Wang, T., Chen, W., Ma, W., et~al.: {LightGBM}: A highly efficient gradient boosting decision tree. In: Advances in Neural Information Processing Systems. vol.~30, pp. 3146--3154 (2017), \url{https://proceedings.neurips.cc/paper_files/paper/2017/hash/6449f44a102fde848669bdd9eb6b76fa-Abstract.html}

\bibitem{ref10}
Lim, B., Ar{\i}k, S.{\"O}., Loeff, N., Pfister, T.: Temporal fusion transformers for interpretable multi-horizon time series forecasting. International Journal of Forecasting  \textbf{37}(4),  1748--1764 (2021). \doi{10.1016/j.ijforecast.2021.03.012}

\bibitem{ref38}
Lin, S., Lin, W., Wu, W., Chen, H., Yang, J.: {SparseTSF}: Modeling long-term time series forecasting with 1k parameters. In: Proceedings of the 41st International Conference on Machine Learning. Proceedings of Machine Learning Research, vol.~235, pp. 30211--30226. PMLR (2024), \url{https://proceedings.mlr.press/v235/lin24n.html}

\bibitem{ref3}
Liu, Z., Cheng, M., Zhao, G., Yang, J., Liu, Q., Chen, E.: Improving time series forecasting via instance-aware post-hoc revision. In: Advances in Neural Information Processing Systems. vol.~38, pp. 40040--40065 (2025). \doi{10.52202/085713-1196}

\bibitem{ref6}
Liu, Z., Yang, Z., Lin, X., Qiu, R., Wei, T., Zhu, Y., et~al.: Breaking silos: Adaptive model fusion unlocks better time series forecasting. In: Proceedings of the 42nd International Conference on Machine Learning. Proceedings of Machine Learning Research, vol.~267, pp. 40022--40042. PMLR (2025), \url{https://proceedings.mlr.press/v267/liu25cm.html}

\bibitem{ref31}
Ma, S., Fildes, R.: Retail sales forecasting with meta-learning. European Journal of Operational Research  \textbf{288}(1),  111--128 (2021). \doi{10.1016/j.ejor.2020.05.038}

\bibitem{ref11}
Makridakis, S., Spiliotis, E., Assimakopoulos, V.: {M5} accuracy competition: Results, findings, and conclusions. International Journal of Forecasting  \textbf{38}(4),  1346--1364 (2022). \doi{10.1016/j.ijforecast.2021.11.013}

\bibitem{ref7}
Mayaluri, Z.L., Mishra, S.: Compositional zero-shot demand forecasting in retail via disentangled demand drivers. International Journal of Forecasting  (2026). \doi{10.1016/j.ijforecast.2026.07.010}, in press

\bibitem{ref35}
{MichalKecera}: Rohlik sales forecasting challenge. Kaggle (2024), \url{https://www.kaggle.com/competitions/rohlik-sales-forecasting-challenge-v2}, dataset. Accessed 20 October 2026

\bibitem{ref30}
Montero-Manso, P., Athanasopoulos, G., Hyndman, R.J., Talagala, T.S.: {FFORMA}: Feature-based forecast model averaging. International Journal of Forecasting  \textbf{36}(1),  86--92 (2020). \doi{10.1016/j.ijforecast.2019.02.011}

\bibitem{ref37}
Nie, Y., Nguyen, N.H., Sinthong, P., Kalagnanam, J.: A time series is worth 64 words: Long-term forecasting with transformers. In: International Conference on Learning Representations (2023), \url{https://openreview.net/forum?id=Jbdc0vTOcol}

\bibitem{ref45}
Ning, K., Pan, Z., Liu, Y., Jiang, Y., Zhang, J., Rasul, K., Schneider, A., Ma, L., Nevmyvaka, Y., Song, D.: {TS-RAG}: Retrieval-augmented generation based time series foundation models are stronger zero-shot forecaster. In: Advances in Neural Information Processing Systems. vol.~38 (2025)

\bibitem{ref43}
Tire, K., Taga, E.O., Ildiz, M.E., Oymak, S.: Retrieval augmented time series forecasting. In: Proceedings of the 29th International Conference on Artificial Intelligence and Statistics. Proceedings of Machine Learning Research, vol.~300, pp. 4294--4302. PMLR (2026), \url{https://proceedings.mlr.press/v300/tire26a.html}

\bibitem{ref32}
Wang, X., Hyndman, R.J., Li, F., Kang, Y.: Forecast combinations: An over 50-year review. International Journal of Forecasting  \textbf{39}(4),  1518--1547 (2023). \doi{10.1016/j.ijforecast.2022.11.005}

\bibitem{ref36}
Zeng, A., Chen, M., Zhang, L., Xu, Q.: Are transformers effective for time series forecasting? Proceedings of the AAAI Conference on Artificial Intelligence  \textbf{37}(9),  11121--11128 (2023). \doi{10.1609/aaai.v37i9.26317}

\bibitem{ref1}
Zhou, S., Sch{\"o}ner, H., Wu, Z., Fouch{\'e}, E., Wilson, I., Wang, S.: Stationarity-aware retrieval-augmented time series forecasting. In: Proceedings of the 32nd ACM SIGKDD Conference on Knowledge Discovery and Data Mining V.2. pp. 6875--6886. Association for Computing Machinery (2026). \doi{10.1145/3770855.3817813}

\end{thebibliography}
\end{document}